# Determination of language families using deep learning, v. 2

P. B. Lerner[1]

Abstract

We use a c-GAN (convolutional generative adversarial) neural network to analyze transliterated text fragments of extant, dead comprehensible, and one dead non-deciphered (Cypro-Minoan) language to establish linguistic affinities. The proposed approach is based on forming a two-dimensional digital "fingerprint" of the fragment. The analysis of the fingerprint employs the methods developed for the study of visual images. Compared to the v.1 of the paper, fingerprint correlations were statistically reliable. At the post-processing stage, we classified the languages through quasi-distances naturally emerging between original and simulated fingerprints by probabilities coming from logistic regression. This paper is agnostic concerning translation and/or deciphering. However, there is hope that the proposed approach can be useful for decipherment with more sophisticated neural network techniques.

## 1. Introduction

The availability of the deep learning instruments made possible the types of analysis unachievable with other methods. Deep learning is used currently in the LLMs (Large Language Models), for image identification, the creation of deepfakes, and analyses of astrophysical and financial information (Krizhevsky, 2012), (Sutskever, 2014), (Vaswani, 2017), (Wang, 2015), (George, 2018), (Li, 2010).

When the instruments of deep learning became widely available, it was decided that the decipherment of all dead languages was only a matter of time. However, the progress is slow and highly application-specific (see the excellent review (Sommershield, 2023) and *op.cit*).

[1] Unaffiliated, Ithaca, NY. Contact: pblerner18@gmail.com, peter.lerner@faculty.umgc.edu.

Deciphered dead languages are still the ones that have bilingual or trilingual texts available with minor exceptions.[2]

This paper implements the c-GAN-based methodology, earlier published by this author in the context of the analysis of causality relationships of the financial time series for the establishment of language families. The significance of this method is that comprehension of the text fragments plays no role in the determination of their phonetic and grammar affinity. The approach can be equally applied to the deciphered and undeciphered languages if only the texts can be converted to a uniform framework. The author uses texts in their Latin transliteration, yet this simply reflects the limitations of the author's resources. In principle, any fragment can be analyzed as long as the signs admit a uniform mapping between them.

Adversarial generative networks (GANs) are found on the following principle. The algorithm consists of two parts: generator and discriminator (critic). The generative part of the network produces arrays from a random noise, which must satisfy some criteria of affinity, usually based on entropic considerations, to the training dataset. The critic must distinguish between a real dataset and its random noise replication by adjusting a large number of parameters (several hundred thousand in my home PC-based simulation and up to trillions for LLMs of Google, etc.). The c-GAN and other neural networks (NNs) allow a significant plethora of computational methods developed with great success for the analysis of visual information.

The author, in his previous analysis of the causality of the financial time series, proposed the following protocol (Lerner, 2023). One chooses the datasets to compare as the training and test datasets, respectively. An entropic comparison measure determines whether a fake, built from random noise by a generator of the neural network, fits better one or the other dataset. In the case of the financial time series, a dataset, that requires more information to fake it from a random noise, is considered more informative. In the case when both datasets are synchronous, a more informative dataset is declared causally determining a less informative dataset.

This protocol in the case of text fragments can be intuitively explained as follows. Imagine a speaker of e.g. Finnish language being asked to produce a series of nonsensical texts claimed to be English. An English-language speaker must determine whether it is nonsense or broken English.

---

[2] S. A. Starostin, private communication to the author, c. 1987. The only exception known to the author was Ventris' decipherment of Linear B, which was, after all, an archaic version of a well-known language (Chadwick, 1990).

Adversely, an English-language speaker is asked to produce a series of nonsensical Finnish fragments, and they are analyzed by a Finn.

If the statistical asymmetry between recognition capabilities is large, we consider the languages linguistically unrelated and vice versa. The more sophisticated the fakes are, the better is an ostensible determination of linguistic affinity.

The paper is structured as follows. In Section 2 we explain the construction of the samples. In Section 3, we provide measures of the distance between datasets. We discuss post-processing metrics in Section 4. The measure of asymmetry between datasets built from the fragments of different languages is used for our analysis in Section 5. The test of robustness is described in Section 6. Section 7 concludes the presentation.

## 2. **Choice of the samples**

We use samples of eight languages. The samples were chosen deliberately small as to 1) be amenable for analysis on a standard PC and 2) conform to a very small existing corps of Cypro-Minoan texts, the only undeciphered language among the eight. Our dataset contains two extant Indo-European (IE) languages (English and Spanish), two extant non-IE (NIE) languages geographically unrelated to the Mediterranean-Fertile Crescent area (Tagalog and Finnish; for simplicity of transliteration), one dead IE (Luwian), and two classified non-IE Middle Eastern languages from the geographic area of Cypro-Minoan (Babylonian—Semitic, and Hurrian—a language isolate) and one unclassified and undeciphered dead language—Cypro-Minoan. After we train the postprocessing algorithm on the six above languages, we classify "out-of-sample" Finnish as a test. Finally, we use the tested classification procedure on a sample of Cypro-Minoan.

Luwian was selected because it was the geographically closest IE area, which was not archaic Greek. Hurrian was selected for the following reasons. The historic period associated with the Cypro-Minoan inscriptions (1500-1150 BCE) almost exactly coincided with the Kassite domination of Mesopotamia. Kassite language is not decisively attributed to any language family, but some researchers related it to Hurro-Urartian (Schneider, 2003), (Fournet, 2011).

In our study, we do not touch on the problem of whether Cypro-Minoan inscriptions were written in one, or several languages (see, e.g. (Palaima, 1989)). Our chosen sample of Cypro-Minoan is too limited to address this possibility.

The choice of Tagalog and Finnish was made because of the original Latin-based writing system simplifying conversion into number lists. The English sample was selected from a finance paper (reviewed by the author) written by the Spanish-speaking authors to complicate recognition. The Spanish sample was taken from the Wikipedia article 'Peres Galdos'. The Tagalog text was chosen from a Tagalog Wikipedia article on the Philippines. Further data on the sources is provided in Table 1.

**Table 1**

| Language | Primary source | Content/article | Size (ns/sp)* |
|---|---|---|---|
| **English** | Draft manuscript | - | 8,356/9,936 |
| **Spanish** | Wikipedia | Peres Haldos | 8,019/9,563 |
| **Tagalog** | Wikipedia | Philippines | 8,441/9,937 |
| **Finnish** | Wikipedia | Kalevala | 11,714/13,187 |
| **Luwian** | (Cambel, 2022) | A prayer to relieve toothache | 9,065/10,387 |
| **Babylonian** | (Clay, 1910) | Random selection | 5,563/6,228 |
| **Hurrian** | (Wegner, 2020), (Campbell, 2016) | Random selection | 5,255/6,219 |
| **Cypro-Minoan**** | (Valerio, 2016) | University of Barcelona Dissertation | 2,070/2,371 |

*ns/sp – size without (ns) and with (sp) spaces in digits.
**All-numbers transliteration

At the preprocessing stage, we fit our transliterated texts into 4×64×64 arrays where text was split into four excerpts to be comparable in size with (usually shorter) excerpts from the dead languages. For preliminary investigation, a standard UTF-8 conversion routine was used.

A standard deep learning procedure includes training, tests, and predictive stages (Aggrawal, 2023). Sometimes, the data undergoes pre-and/or post-processing for analysis. This is, in part, because the multidimensional tensors, which are frequently the numeric output of the neural nets, are difficult to convey in the usual two-dimensional format of a text or an image. Our pre-processing stage included digitizing the input texts and fitting individual excerpts into a standard 64×64=4196 signs list, for which some entries could be empty. This list has a visual form, which the author called a "fingerprint" in a financial context (Lerner, 2023). A fingerprint is a unique image, which we prepared to analyze through the c-GAN network. We choose one of the 4-tuples of a given language fingerprint as training and another 4-tuple as a test sample.

C-GAN network is run to produce fake fingerprints from a random noise using a training file, and then analyzes it with a discriminator/critic. The outputs of the neural network were assigned certain affinity criteria (quasi-distances, see the next section), which we used to ascribe hypothetical relationships between languages or a lack thereof.

## 3. **Distances on the state space**

Because of the "black box" nature of the neural networks, their output is hard to rationalize. First, the human mind has evolved to analyze two- or three-dimensional images in three-dimensional space. Most humans cannot directly comprehend tensor inputs, intermediate results, and outputs typical for neural networks. Second, the results of neural network analyses are necessarily stochastic and depend on the large number of estimated intrinsic parameters, which are frequently inaccessible, but in any case, too numerous for humans to rationalize. Third, deep learning results can depend on how the training and testing samples are organized, even if they represent identical datasets. All of this can indicate the failure of a deep learning procedure, but it can also show additional information we fail to recognize (Brownlee, 2021). Neural networks are the "black boxes", so instead of the interpretation of a hundred thousand—in my case, and trillions of—parameters in the case of Google and Microsoft Large Language Models, one must design numerical experiments and analyze the output from a deep learning algorithm in its entirety.

To systematize the results, we propose two measures of divergence of images as follows: After the c-GAN generated fake images ("fingerprints") of the session, we considered these images as (1) matrices, and (2) nonnormalized probability distributions.

The first approach is to treat arrays as matrices (tensors). We computed the pseudo-metric cosine between the image arrays $X$ and $Y$ according to the following formula:

$$C_{XY} = \frac{\|X+Y\|^2 - \|X-Y\|^2}{4\|X\|\cdot\|Y\|} \qquad (1)$$

In the above formula, the norm $||\cdot||$ is a Frobenius matrix norm representing each image array. In the first stage, we computed the distance as the average of each twentieth or sixteenth of the last 400 images in the sequence. Because, sometimes, the fake image is an empty list having a zero norm, we modified this formula according to the following prescription:

$$C_{train,fake} = \frac{\|train + fake\|^2 - \|train - fake\|^2}{4\|train\| \cdot \|test\|} \tag{2}$$

$$C_{test,fake} = \frac{\|test + fake\|^2 - \|test - fake\|^2}{4\|train\| \cdot \|test\|}$$

Equation (2) provides answers close to the correct geometric formula (2), but it does not fail in the case of an empty fake image. The pseudo-metric measure, calculated according to Equation (2), provides a fair picture of the affinity of the fake visual images to the originals, but it is still unstable with respect to the different stochastic realizations of the simulated images.

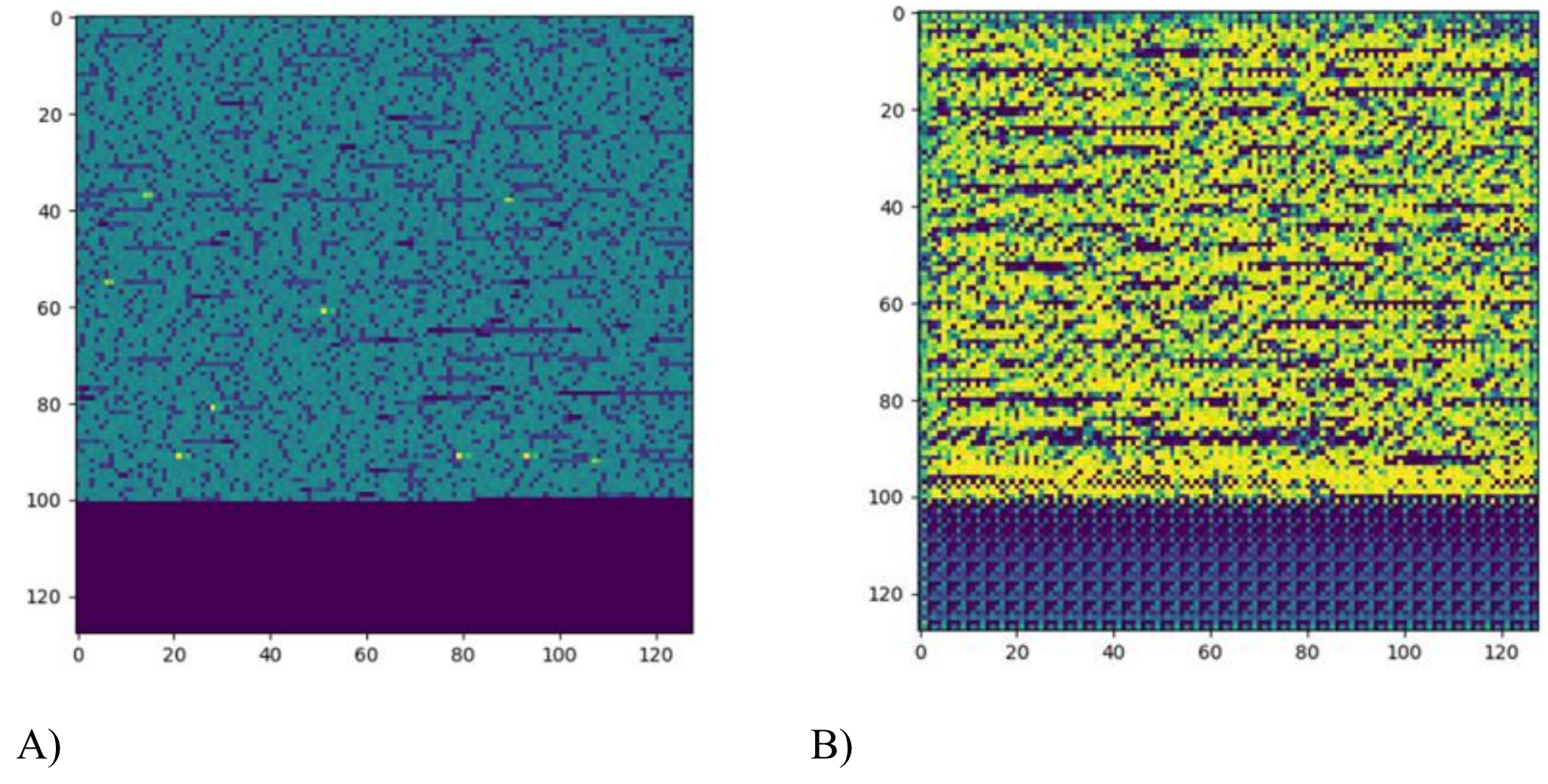

Fig. 2. A) Fingerprint of an original English 128x128 data file. It splits into four 64×64 training files. A dark band indicates the end of the sample. B) Fingerprint of a Spanish text. The visual difference with the English fingerprint is evident.

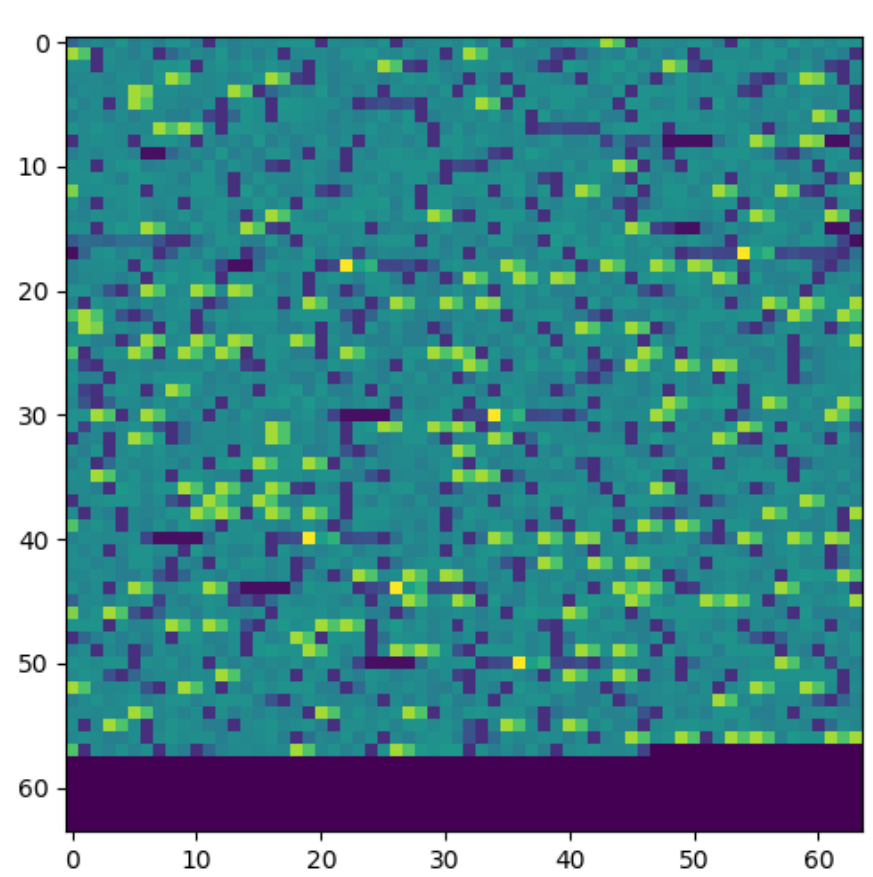

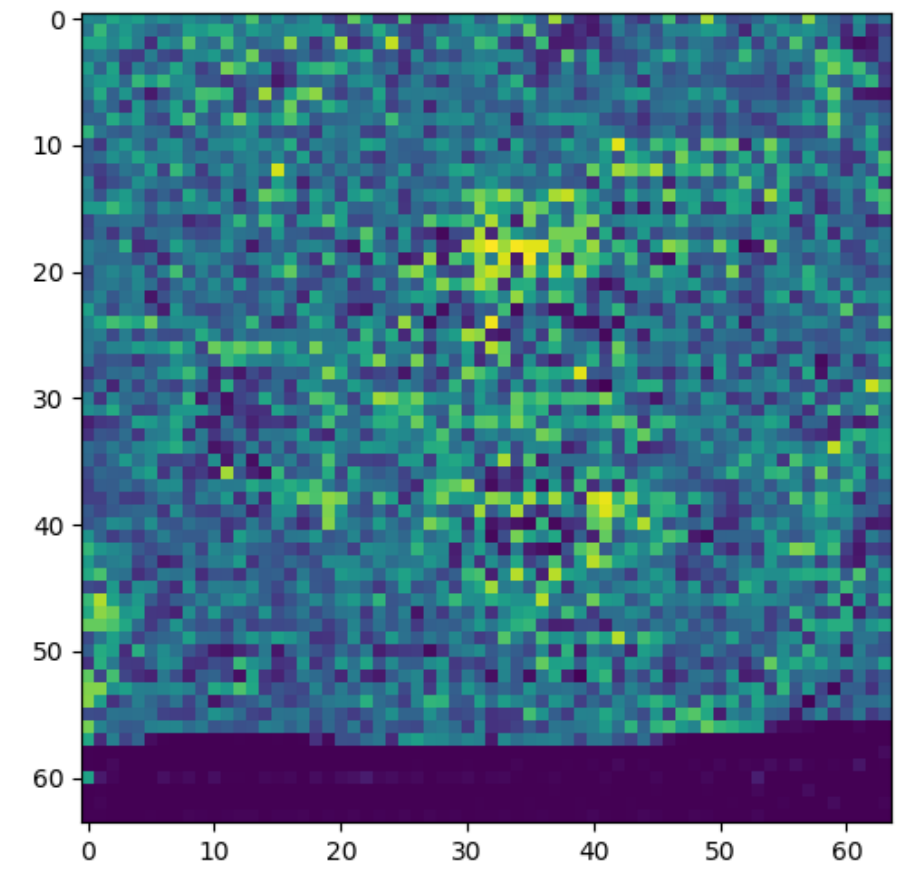

Fig. 3. The c-GAN output for a Finnish fingerprint and a "good" Finnish fake (~30% correlation) tested on a Tagalog sample.

4. **Post-processing**

To analyze affinity between languages, we need some sense of distance between fake images, i.e. the arrays by which the c-GAN simulates relations between symbols of an original text. The results of our simulation are given in Table 1.

Table 1

| | First | Second | C_1 | C_2 |
|---|---|---|---|---|
| 1 | Hurr | Luw | 0.0055 | 0.048 |
| 2 | Luw | Hurr | 0.3357 | 0.2747 |
| 3 | Hurr | Bab | 0.00053 | 0.00055 |
| 4 | Bab | Hurr | 0.2453 | 0.2769 |
| 5 | Sp | Tag | 0.0015 | 0.0015 |
| 6 | Tag | Sp | 0.4129 | 0.4122 |
| 7 | Sp | Luw | 0.085 | 0.0772 |
| 8 | Luw | Sp | 0.01371 | 0.01315 |
| 9 | Luw | Bab | 0.3594 | 0.4354 |
| 10 | Bab | Luw | 0.17 | 0.1678 |
| 11 | Sp | Bab | 0.4444 | 0.4354 |
| 12 | Bab | Sp | 0.0133 | 0.0102 |
| 13 | Tag | Luw | 0.05333 | 0.05907 |
| 14 | Luw | Tag | 0.3027 | 0.3037 |
| 15 | Eng | Tag | 0.3659 | 0.375 |
| 16 | Tag | Eng | 0.3583 | 0.3152 |
| 17 | Eng | Sp | 0.3266 | 0.326 |
| 18 | Sp | Eng | 0.00811 | 0.0076 |
| 19 | Eng | Bab | 0.1709 | 0.1641 |
| 20 | Bab | Eng | 0.01213 | 0.00993 |
| 21 | Eng | Hurr | 0.2888 | 0.2442 |
| 22 | Hurr | Eng | 0.00046 | 0.00033 |
| 23 | Eng | Suo | 0.005 | 0.0054 |
| 24 | Suo | Eng | 0.3436 | 0.2365 |
| 25 | Sp | Suo | 0.2882 | 0.2689 |
| 26 | Suo | Sp | 0.00056 | 0.00051 |
| 27 | Tag | Suo | 0.3659 | 0.3917 |
| 28 | Suo | Tag | 0.3925 | 0.2788 |
| 29 | Suo | Luw | 0.2954 | 0.2388 |
| 30 | Luw | Suo | 0.00055 | 0.00043 |
| 31 | Suo | Hurr | 0.4501 | 0.1901 |
| 32 | Hurr | Suo | 0.2604 | 0.2413 |
| 33 | Suo | Bab | 0.4123 | 0.3562 |
| 34 | Bab | Suo | 0.00052 | 0.00056 |
| 35 | Eng | Luw | 0.2891 | 0.2891 |
| 36 | Luw | Eng | 0.00029 | 0.0003 |
| 37 | Sp | Hurr | 0.3221 | 0.2558 |
| 38 | Hurr | Sp | 0.00023 | 0.0002 |
| 39 | Tag | Hurr | 0.2975 | 0.2169 |
| 40 | Hurr | Tag | 0.3599 | 0.3465 |
| 41 | Tag | Bab | 0.359 | 0.3465 |
| 42 | Bab | Tag | 0.2237 | 0.129 |
| 43 | Minoa | Eng | 6.73E-05 | 7.54E-05 |
| 44 | Eng | Minoa | 0.0013 | 0.0015 |
| 45 | Sp | Minoa | 0.5186 | 0.1776 |
| 46 | Minoa | Sp | 0.1839 | 0.1437 |
| 47 | Tag | Minoa | 0.6157 | 0.159 |
| 48 | Minoa | Tag | 0.1785 | 0.1529 |
| 49 | Luw | Minoa | 0.4519 | 0.1328 |
| 50 | Minoa | Luw | 0.1807 | 0.1088 |
| 51 | Hurr | Minoa | 0.000122 | 0.000091 |
| 52 | Minoa | Hurr | 0.2403 | 0.334 |
| 53 | Bab | Minoa | 0.5303 | 0.1966 |
| 54 | Minoa | Bab | 0.1736 | 0.1479 |
| 55 | Suo | Minoa | 0.5396 | 0.0918 |
| 56 | Minoa | Suo | 0.2264 | 0.347 |

***Note**. Tested languages are grouped in 28 pairs according to train-test relationships in the c-GAN network. The lowest correlation between the original fingerprints and their fakes is 6.73e-05 and the highest correlation is 0.5396. C_1 is the correlation between the original fingerprint and the training fake, C_2 is the correlation between the original fingerprint and the test fake.

By the method of our data representation, each pair of languages is described by 2×2 matrix. For our preliminary classification, we use two invariants of this matrix: trace and (absolute value) of the determinant as classifiers as in Example.

**Example**

| First | Second | C_1 | C_2 | det | Tr |
|---|---|---|---|---|---|
| Hurr | Luw | 0.0055 | 0.048 | 0.0146 | 0.2802 |
| Luw | Hurr | 0.3357 | 0.2747 | | |

An initial sample without Finnish and Minoan (15 pairs) was used to train the Mathematica© classifier algorithm with the logistic regression method. Classification categories are given as 'A' – no relationship, 'B' – possible geographic or contemporaneous relationship, 'C' – pair is linguistically related, and 'D' – pair is linguistically related and contemporaneous (only English and Spanish in our sample). We check our classification on an “Out-of-Sample” Finnish sample (6 additional pairs of languages). The results are given in Table 2. We observe that our testing procedure rules out a linguistic affinity for five pairs out of six – except for the Spanish-Finnish pair – for which the acceptance statistic is poor.

**Table 2** “Out-of-Sample” Finnish is classified. Only its pairing with Spanish is misclassified, albeit with a small probability.

| **Language pair** | **Classification probability** |
|---|---|
| **Eng-Suo** | A→0.4232, B→0.2471, C→0.2016, D→0.1281 |
| **Sp-Suo** | C→0.3138, B→0.2891, A→0.2822, D→0.1149 |
| **Tag-Suo** | A→0.7473, B→0.1781 |
| **Luw-Suo** | A→0.2972, C→0.2952, B→0.2915, D→0.1161 |
| **Hurr-Suo** | A→0.7645, B→0.1700 |
| **Bab-Suo** | A→0.4735, B→0.2948, C→0.1194, D→0.1124 |

We consider these results rather remarkable given that 1) the neural networks do not know anything about linguistics, and 2) the author is not a professional linguist.

**Table 3**. Classification of the Cypro-Minoan sample.

| Language pair | Classification probability |
|---|---|
| **Eng-Minoa** | C→0.7658, B→0.1146 |
| **Sp-Minoa** | A→0.7496, B→0.1783 |
| **Tag-Minoa** | A→0.7884, B→0.1581 |
| **Luw-Minoa** | A→0.7037, B→0.1985, D→0.0769 |
| **Hurr-Minoa** | A→0.3873, B→0.2934, C→0.1999, D→0.1195 |
| **Bab-Minoa** | A→0.7557, B→0.1754 |
| **Suo-Minoa** | A→0.8297, B→0.1323 |

Results of the classification of the Minoan sample by the same tree algorithm with logistic regression rule out Minoan affinity with any language except the English. This random fact can possibly be attributed to the particular defects of transliteration. The negation statistic of Minoan-Hurrian pair is low. We hope that our further investigation clarifies these details.

## 7. Conclusion

In this paper, the author proposes a deep learning-based approach to language classification. This approach uses the comparison of digital fingerprints (see Section 2) of language samples. We define a set of quasi-distances on the state space to quantify the affinity between languages in our sample. This method does not require that the classified text be comprehensible or even transcribed, as long as signs/characters in it are uniformly digitized. Despite a failure in classification – the results of v. 1 are not supported when the correlation between fingerprints was improved, we express hope that more advanced applications of his approach can bring Cypro-Minoan inscriptions closer to decipherment (Ferrara, 2019).